\documentclass{article}

\usepackage{PRIMEarxiv}

\usepackage[utf8]{inputenc} 
\usepackage[T1]{fontenc}    
\usepackage{hyperref}       
\usepackage{url}            
\usepackage{booktabs}       
\usepackage{amsfonts}       
\usepackage{nicefrac}       
\usepackage{microtype}      
\usepackage{lipsum}
\usepackage{fancyhdr}       
\usepackage{graphicx}  
\usepackage{wrapfig}
\usepackage[table]{xcolor}
\graphicspath{{media/}}     
\usepackage{amsmath}
\usepackage{pifont}
\usepackage{amssymb}

\title{Multimodal Masked Autoencoder Pre-training for 3D MRI-Based Brain Tumor Analysis with Missing Modalities

}

\author{
  Lucas Robinet \\
  Oncopole Claudius Régaud \\
  IRT Saint Exupéry \\
  INSERM Cancer Research Center of Toulouse \\
  Toulouse\\
  \texttt{robinet.lucas@iuct-oncopole.fr} \\
   \And
  Ahmad Berjaoui \\
  IRT Saint Exupéry \\
  INSERM Cancer Research Center of Toulouse \\
  Toulouse\\
  \texttt{ahmad.berjaoui@irt-saintexupery.com} \\
  \And
  Elizabeth Cohen-Jonathan Moyal  \\
  Oncopole Claudius Régaud \\
  INSERM Cancer Research Center of Toulouse \\
  Toulouse\\
  \texttt{moyal.elisabeth@iuct-oncopole.fr} \\
  }

\begin{document}

\maketitle

\begin{abstract}
Multimodal magnetic resonance imaging (MRI) constitutes the first line of investigation for clinicians in the care of brain tumors, providing crucial insights for surgery planning, treatment monitoring, and biomarker identification.
Pre-training on large datasets have been shown to help models learn transferable representations and adapt with minimal labeled data.
This behavior is especially valuable in medical imaging, where annotations are often scarce.
However, applying this paradigm to multimodal medical data introduces a challenge: most existing approaches assume that all imaging modalities are available during both pre-training and fine-tuning.
In practice, missing modalities often occur due to acquisition issues, specialist unavailability, or specific experimental designs on small in-house datasets.
Consequently, a common approach involves training a separate model for each desired modality combination, making the process both resource-intensive and impractical for clinical use.
Therefore, we introduce BM-MAE, a masked image modeling pre-training strategy tailored for multimodal MRI data. 
The same pre-trained model seamlessly adapts to any combination of available modalities, extracting rich representations that capture both intra- and inter-modal information.
This allows fine-tuning on any subset of modalities without requiring architectural changes, while still benefiting from a model pre-trained on the full set of modalities.
Extensive experiments show that the proposed pre-training strategy outperforms or remains competitive with baselines that require separate pre-training for each modality subset, while substantially surpassing training from scratch on several downstream tasks.
Additionally, it can quickly and efficiently reconstruct missing modalities, highlighting its practical value.
Code and trained models are available at: \href{https://github.com/Lucas-rbnt/BM-MAE}{https://github.com/Lucas-rbnt/BM-MAE}
\end{abstract}

\section{Introduction}
\begin{figure}[!ht]
\centering
\includegraphics[scale=0.79]{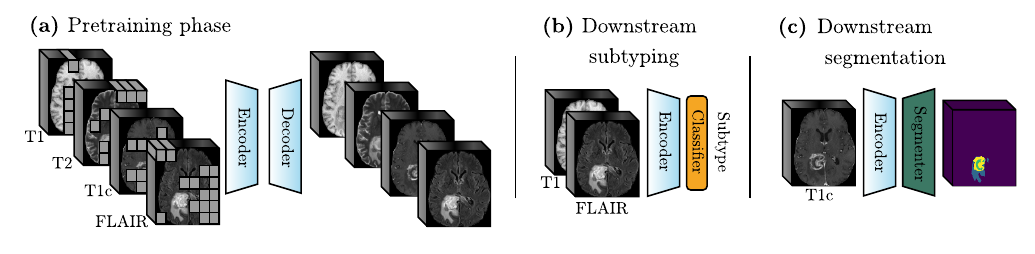}
\caption{(a) BM-MAE pre-training on all modalities through the reconstruction of masked patches. The trained encoder can handle any modality combination. (b) Fine-tuning the encoder for subtyping on T1 and FLAIR. (c) Fine-tuning the same encoder for segmentation on T1c.}
\label{fig:intro}
\end{figure}
Gliomas are the most common type of primary brain tumors, originating from glial cells and often leading to poor patient prognosis due to their potential rapid growth and infiltration into surrounding brain tissue \cite{schwartzbaum_epidemiology_2006}.
These tumors comprise complex subtypes that pose significant treatment challenges, driven by diverse genetic profiles and variable responses to therapy, complicating both surgical and therapeutic interventions \cite{chen_glioma_2017,louis_2021_2021}.
In this critical context, multimodal anatomical MRI plays a central role in the clinical workflow for brain tumor assessment.
Hence, standard sequences, including T1-weighted (T1), contrast-enhanced T1-weighted (T1c), T2-weighted (T2), and Fluid Attenuated Inversion Recovery (FLAIR), provide complementary information essential for accurate diagnosis, surgical planning, and treatment monitoring \cite{henson_mri_2005}.

Recent advances in self-supervised learning (SSL) have opened new possibilities for learning rich representations from unlabeled data through the use of pretext tasks.
Two prominent self-supervised learning paradigms have gained widespread adoption. 
The first is contrastive learning, exemplified by SimCLR \cite{pmlr-v119-chen20j}, which aligns representations of different views of the same input.
The second is masked image modeling, as used in Masked Autoencoders (MAE) \cite{he_masked_2022}, where the model learns to reconstruct missing regions from a visible context.
These strategies have been increasingly adopted in medical imaging to alleviate the reliance on large annotated datasets.
In brain MRI, adaptations of SimCLR \cite{ali_self-supervised_2021} and MAE \cite{10230477} have shown strong performance on downstream segmentation tasks, allowing models to learn meaningful features directly from raw scans.
However, an important limitation of these existing SSL methods is their assumption of full modality availability during both pre-training and fine-tuning.
Typically, MRI modalities are stacked along the channel axis, enforcing a fixed input shape.
This design fails to accommodate the reality of clinical datasets, which often suffer from missing modalities due to scanner availability, acquisition time constraints, or inconsistent imaging protocols.
While some approaches attempt to address this issue, they are usually tailored for segmentation and depend on task-specific strategies, including imputation or multi-encoder architectures, to handle incomplete inputs \cite{havaei_hemis_2016,hu_knowledge_2020,azad_smu-net_2022,shen_brain_2019,wang_acn_2021,zhang_mmformer_2022,liu_m3ae_2023}.
As a result, a generalizable multimodal pre-training strategy capable of handling missing modalities remains an open challenge.
In the absence of such an approach, pre-training must be repeated for each combination of available modalities, making it computationally expensive and hindering knowledge transfer between modalities.
This has led to two main practices in clinical studies involving deep learning: some attempt to pre-train models on the specific subset of modalities they are using \cite{robinet_mri-based_2023}, a process that is tedious and cumbersome; others, train models from scratch \cite{zeineldin_deepseg_2020,tupe-waghmare_comprehensive_2021}, which may result in suboptimal performance.

Inspired by the growing popularity of Multimodal MAEs in computer vision \cite{bachmann_multimae_2022,geng2022multimodalmaskedautoencoderslearn,4m}, we introduce Brain Multimodal MAE (BM-MAE) to address these aforementioned challenges.
BM-MAE is a scalable, attention-based pre-training strategy designed to learn rich cross-modal representations while efficiently handling the complexity of high-dimensional 3D multimodal data.
A unique Vision Transformer (ViT) encoder \cite{dosovitskiy_image_2021} is first pre-trained on multimodal data through masked image modeling, where large portions of the input are masked and the model learns to reconstruct them. 
After the pre-training phase, the encoder can be adapted to any modality combinations and be fine-tuned on diverse downstream tasks, as illustrated in Figure~\ref{fig:intro}.
To the best of our knowledge, this is also the first work proposing to handle missing modalities in multimodal brain MRI using a single shared ViT.
To thoroughly assess the effectiveness and flexibility of our pre-training strategy, we conduct a comprehensive set of experiments across multiple clinical tasks and modality configurations.
Hence, first an encoder is pre-trained on the BraTS2021 \cite{baid_rsna-asnr-miccai_2021} dataset using all four anatomical modalities.
Then, this same encoder is fine-tuned across every modality combinations for various downstream tasks including glioma segmentation, glioma subtyping and survival prediction. 
Across all evaluated tasks and modality configurations, fine-tuning the pre-trained encoder consistently leads to strong and reliable performance, demonstrating both its flexibility and robustness.

\section{Related Work}
Across the following paragraphs, we discuss two main areas of related work. 
The first focuses on established self-supervised learning methods in medical imaging, which aim to learn transferable representations from unlabeled data. 
Their final objective is similar to that of BM-MAE.
The second covers approaches developed to handle missing modalities in the context of segmentation. 
Although these methods pursue a different goal, they provide valuable insights into architectural and methodological strategies for managing incomplete multimodal data.

\subsection{Self-Supervised Learning for Medical Image Analysis}
Self-supervised learning is gaining momentum in medical imaging, with promising approaches emerging to leverage unlabeled data. 
For instance, Ouyang et al. \cite{ouyang2021representationdisentanglementmultimodalbrain} propose a method based on 2D slices of MRI and PET to disentangle shared and modality-specific representations.
Following a more conventional contrastive paradigm, Ali et al. \cite{ali_self-supervised_2021} propose using SimCLR to pre-train an encoder on brain MRI scans. 
Two views of the same patient are generated through data augmentations. 
The model is then trained to encourage high cosine similarity between views of the same input while minimizing similarity with other samples in the batch through an InfoNCE loss \cite{oord}.
While this approach yields promising results, its effectiveness heavily depends on the choice of data augmentations \cite{gowda2024breakfreestrongdata}. 
In medical imaging, however, standard augmentations such as intensity shifts, random cropping, and rotations are often harder to apply consistently due to anatomical constraints and clinical relevance. 
Additionally, contrastive methods are sensitive to batch size, as larger batches provide more accurate estimates of mutual information \cite{oord}. 
This poses a challenge in 3D settings, where memory constraints typically limit batch size. 
Zhou et al. \cite{10230477} address this by using MAE pre-training, which reduces reliance on both data augmentation and large batches. 
Their approach involves randomly masking a substantial number of patches from the input MRI volumes and training the model to reconstruct the missing content.
These methods do not inherently support missing modalities, and the pre-training process must be repeated for each desired combination of input modalities. In contrast, BM-MAE offers a unified strategy that handles all modality combinations from a single pre-training.

\subsection{Multimodal Medical Image Analysis with Missing Modalities}
Although it differs from our objective of self-supervised learning, segmentation with missing modalities has been more extensively studied.
Methods generally fall into two categories: those that also require separate training for each modality combination, often relying on knowledge distillation \cite{hu_knowledge_2020} or adversarial objectives \cite{wang_acn_2021,azad_smu-net_2022}, and those that aim to handle all missing scenarios with a single model \cite{dorent_hetero-modal_2019, zhang_mmformer_2022,liu_m3ae_2023}.
Among the latter, \cite{dorent_hetero-modal_2019} propose the use of multimodal variational autoencoders to reconstruct missing modalities during segmentation.
Their approach uses modality-specific encoders to map each observed modality into a shared latent space.
Liu et al. \cite{liu_m3ae_2023} introduce a two-stage UNet-based framework for segmentation under missing modalities. 
Their approach first pre-trains the model with random modality dropout and patch masking, then fine-tunes it through a self-distillation process. 
Missing modalities are handled via model inversion producing a single, dataset-specific, substitute for missing inputs, which may hinder generalization and interpretability.
Architecturally, the work most closely related to ours is likely mmFormer \cite{zhang_mmformer_2022}, which uses modality-specific transformer-based encoders followed by an attention module to fuse the resulting tokens and handle missing modalities.
The fused representation tokens are then passed to a convolutional decoder to output a semantic segmentation map.
Beyond the difference in objectives —segmentation versus self-supervised learning— BM-MAE clearly differs architecturally in key ways: it uses a single transformer encoder shared across all modalities (rather than one per modality), a decoder composed entirely of transformer layers, and a masked input strategy that processes only a subset of image patches, resulting in improved computational efficiency.

\section{Brain Multimodal Masked Autoencoder}
Let \( M = \{T_1, T_{1c}, T_2, {FLAIR}\} \) be the set of MRI modalities, and let \(\mathcal{X} = \{X_m\}_{m \in M}\) denote the multimodal MRI data for a given patient. 
Hence, each modality \( X_m \in \mathbb{R}^{H \times W \times D} \) represents a 3D volumetric image, where \( H \), \( W \), and \( D \) denote the height, width, and depth of the volume, respectively.
 
\paragraph{Patching and positional embeddings.} Each \(X_m\) is reshaped into a sequence of patches \(X_m^p \in \mathbb{R}^{L \times p^3}\) with \(p\) denoting patch size and \(L=H\cdot W\cdot D/p^3\) the number of patches. 
Note that here, since all modalities share the same spatial dimensions, we use a common patch size \(p\), resulting in the same number of patches \(L\) per modality.
Afterwards, each modality \( m \) is passed through a modality-specific linear projection \(l_m: \mathbb{R}^{p^3} \mapsto \mathbb{R}^{d}\), which maps input patches into a shared embedding space, producing modality-specific tokens.
Spatial information is then incorporated by adding a fixed sine-cosine positional encoding \( \mathrm{PE} \in \mathbb{R}^{L \times d} \) to to the embedded patches of each modality.
Formally, the resulting token sequence for modality \( m \) is given by \( S_m = l_m(X_m^p) + \mathrm{PE} \in \mathbb{R}^{L \times d} \).

\paragraph{Masking strategy.} As mentionned in MAE \cite{he_masked_2022}, the mask sampling technique can significantly influence performance on downstream tasks. 
Following prior work in traditional computer vision \cite{bachmann_multimae_2022,4m}, we define a global masking ratio \( r \in [0,1] \), that controls the number of visible tokens, given by
\(
L^v = \left\lfloor (1 - r) \cdot |M| \cdot L \right\rfloor,
\)
where \(|M| \cdot L\) is the total number of tokens across all modalities.  
To distribute visible tokens across modalities, we sample modality-specific proportions \((w_1, \ldots, w_{|M|})\) from a symmetric Dirichlet distribution, \((w_1, \ldots, w_{|M|}) \sim \text{Dir}(\boldsymbol{\alpha})\), satisfying \( \sum_{m=1}^{|M|} w_m = 1 \).  
Given the total number of visible tokens \( L^v \), each modality \( m \) is allocated \( L_m^v = \lfloor w_m \cdot L^v \rfloor \) tokens, selected uniformly at random from its patch sequence. 
The resulting sequence of visible tokens for modality \( m \) is denoted by \( \overline{S}_m \in \mathbb{R}^{L_m^v \times d} \).
As discussed in \cite{bachmann_multimae_2022}, Dirichlet-based sampling induces diverse masking patterns by allowing some modalities to be almost entirely masked, promoting representational diversity.
\newline
\begin{figure*}[t]
\centering
\resizebox{\textwidth}{!}{
\includegraphics[scale=0.7]{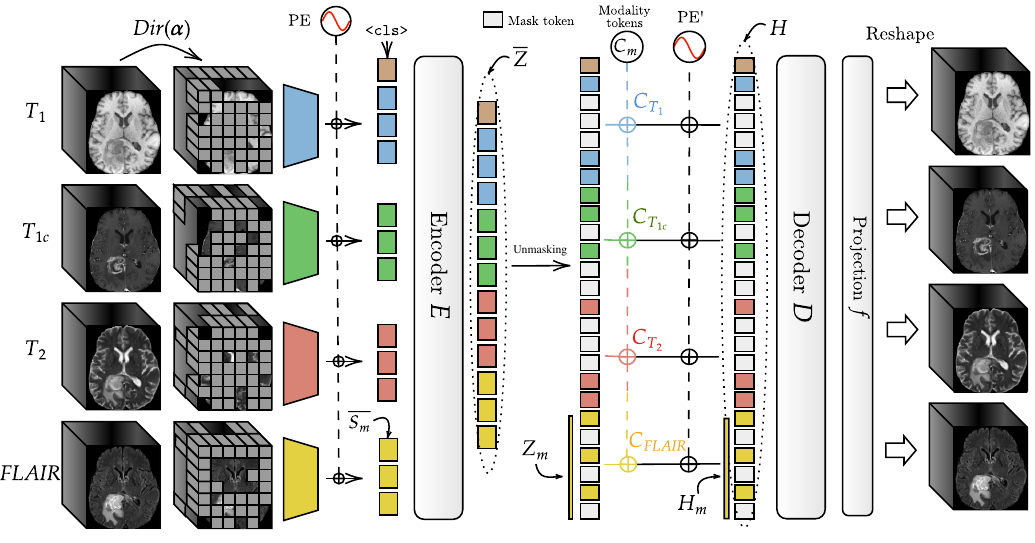}
}
\caption{Overview of the proposed BM-MAE architecture. 
A detailed view of the decoding mechanism with modality and mask tokens is provided. 
For clarity, token notations (e.g \(\overline{S_m}, Z_m, H_m\)) are shown for a single modality, though the same applies to all.}
\label{fig:framework}
\end{figure*}

\paragraph{Multimodal encoder.}
For the encoder input, all \textit{visible} tokens from each modality are concatenated into a single sequence:
\[\overline{S} = \text{Concat}(\texttt{<cls>}, \{\overline{S_m}\}_{m\in M}) \in \mathbb{R}^{(L^v + 1)\times d}\] 
A learnable global \texttt{<cls>} token is prepended to the sequence to aggregate a patient-level representation, resulting in a sequence of length \(L^v + 1 \).
Then, this sequence is given to an encoder \(E\) composed of transformer blocks shared across all modalities, as shown in Figure~\ref{fig:framework}:
\[\overline{Z} = E(\overline{S}) \in \mathbb{R}^{(L^{v} + 1) \times d}\]
This design enables the model to capture both spatial structure within modalities and cross-modal relationships.
Since \(E\) operates solely on visible (i.e., unmasked) patches, the computational cost during training is significantly reduced, as \(L^v < |M|\cdot L\).

\paragraph{Mask and modality tokens.} 
Encoded tokens are projected into a lower-dimensional latent space of size \( d' \) using a learnable linear layer.
This allows to reduce computational overhead as the decoder is only an auxiliary module used for the pre-training phase.
The resulting tokens are then unshuffled, and learnable mask tokens are inserted at the positions corresponding to the originally masked patches. 
We then regroup the resulting tokens by modality, and denote by \( Z_m \in \mathbb{R}^{L \times d'} \) the complete sequence for modality \( m \), consisting of both encoded and mask tokens.
Each \( Z_m\) is augmented by adding a fixed sine-cosine positional encoding \( \mathrm{PE'} \in \mathbb{R}^{L \times d'} \) and a modality-specific learnable embedding \( C_m \in \mathbb{R}^{d'} \), shared across all tokens of the sequence: \( H_m = Z_m + C_m + \mathrm{PE'}\).
This augmentation provides the decoder with both spatial and modality information necessary for accurate patch reconstruction.

\paragraph{Multimodal decoder.} 
The multimodal part follows a similar strategy as the encoder.
First, the modality-augmented sequences \( \{H_m\}_{m\in M}\) are concatenated together with the global \texttt{<cls>} token.
The resulting sequence \( H \in \mathbb{R}^{(|M|\cdot L + 1) \times d'}\) is then passed through a lightweight transformer decoder \(D\) composed of standard self-attention blocks but with fewer layers to reduce computational cost.
Each decoded token is mapped back to the original patch space using a shared linear projection  \(
f: \mathbb{R}^{d'} \rightarrow \mathbb{R}^{p^3}
\).
Finally, predicted patches are reshaped to reconstruct the input MRIs, yielding a reconstructed volume \(\hat{X_m}\) for each modality \( m \).

\paragraph{Loss function.} As in MAE \cite{he_masked_2022}, the model is trained to minimize a mean squared error (MSE) computed only on masked patches. 
Specifically, the loss is defined as
\[
\mathcal{L}_{recons} = \frac{1}{|M|} \sum_{m \in M} \frac{1}{|\mathcal{M}_m|} \sum_{i \in \mathcal{M}_m} \left\| \hat{X}_m^i - X_m^i \right\|^2
\]
where \( \mathcal{M}_m \) denotes the set of masked patch indices for modality \( m \).
Examples of sampled mask alongside the corresponding reconstructions are provided in Supplementary Material Figure~\ref{fig:mask_sampling}.

\section{Experiments}
In this section, we study the relevance of the representations learned by BM-MAE.
We assess the potential of BM-MAE as a general-purpose backbone across a diverse set of clinical downstream tasks. 
Specifically, we investigate: (i) if using BM-MAE pre-trained encoder consistently leads to better performance compared to training from scratch; and (ii) whether pre-training on all modalities helps the encoder capture cross-modal information, leading to better fine-tuning performance across different modality subsets.
We also visualize both the quality of reconstructions and the consistency of the produced embeddings across modality subsets to gain insights into the cross-modal relationships learned by our model.

\subsection{Pre-training details}
\paragraph{Dataset and data preprocessing.} BM-MAE is trained on the widely used BraTS2021 dataset \cite{baid_rsna-asnr-miccai_2021}, which includes the four common anatomical MRI modalities. 
The dataset consists of 1,251 training cases and 219 test cases.
For each patient, all volumes undergo skull-stripping, are co-registered to a common anatomical template, and resampled to a uniform isotropic resolution of \(1mm^3\). 
To ensure consistency, each volume is centrally cropped to a fixed size of \(128 \times 128 \times 128\) and standardized modality-wise.
Pre-processing are implemented using MONAI \cite{cardoso_monai_2022}.

\paragraph{TCGA subset.} 
A subset of 165 patients from this dataset also appears in the TCGA collection \cite{chang_cancer_2013}, for which additional clinical labels are available, including tumor subtype, either lower-grade glioma (LGG) or glioblastoma (GBM), and survival time.
To avoid data leakage during downstream evaluation on classification and survival tasks, we perform an additional pre-training run in which these 165 patients are excluded. 

\paragraph{Comparisons.}
We compare BM-MAE with SimCLR \cite{ali_self-supervised_2021} and a standard MAE \cite{10230477}, all using the same ViT backbone \cite{dosovitskiy_image_2021}.
In SimCLR and standard MAE, a single tokenizer is used, which expects a fixed input shape with modalities stacked as channels.
As a result, we repeat pre-training independently for each modality combination.
In contrast, BM-MAE employs a modality-specific tokenizer for each input modality, followed by a shared encoder, allowing a single unified pre-training phase across all combinations.
In our setup, this results in BM-MAE being pre-trained only once, while SimCLR and standard MAE require 15 independent runs, one per modality combination.

\paragraph{Implementation details.}
The encoder consists in 12 transformer blocks, each with 12 attention heads, an embedding dimension \( d = 768 \), and a multilayer perceptron (MLP) dimension of 1536.
The patch size is set to \(16\times 16\times 16\) voxels. 
Each decoder is composed of 3 transformer blocks with 12 attention heads and an embedding dimension \(d' = 384\). 
We set the concentration parameter \(\boldsymbol{\alpha}\) of the Dirichlet distribution to 1, yielding a uniform prior over masking scenarios.
Following common practice \cite{he_masked_2022}, we use \(r=0.75\) as the masking ratio, resulting in \(N^v=512\) visible tokens for BM-MAE and \(128\) for regular MAE.
Both MAE and BM-MAE are pretrained for 1000 epochs with a batch size of 6, while SimCLR is pretrained for 400 epochs with a batch size of 4 and a temperature parameter \(\tau = 0.05\). 
All pre-training use the AdamW optimizer with a base learning rate of \(1 \times 10^{-4}\) and a weight decay of 0.05. 
Training starts with a 50-epoch warm-up, followed by learning rate decay using a cosine schedule. 
Experiments are conducted on a single NVIDIA GeForce RTX 4090.

\subsection{Downstream Tumor Segmentation} 
\paragraph{Glioma Subregion Segmentation.}
We evaluate BM-MAE on glioma subregion segmentation using BraTS 2021 \cite{baid_rsna-asnr-miccai_2021}, which includes labels for necrotic core, enhancing tumor, and peritumoral edema. 
Following the competitions evaluation rules \cite{baid_rsna-asnr-miccai_2021}, we report Dice scores on three composite classes: whole tumor (WT), enhancing tumor (ET), and tumor core (TC).
Hence, we adopt the self-pretraining paradigm from \cite{10230477} and evaluate BM-MAE fine-tuning on the same dataset.
This is the most challenging setting we confront BM-MAE with. 
This removes any advantage typically gained from transferring to a smaller target set. 
It allows us to directly assess whether this leads to improved segmentation performance, as observed in \cite{10230477}, and whether pre-training on all four modalities enables better generalization when some modalities are missing at fine-tuning, particularly for hard subregions.

\paragraph{Implementation details.} 
Pre-trained ViTs are used as the backbone of a UNETR model \cite{hatamizadeh_unetr_2022}. 
The formulation \textit{Scratch} refers to the training of the same UNETR from a random initialization.
In BM-MAE, each input modality contributes its own set of tokens after the ViT encoder, leading to a variable number of tokens depending on the number of available modalities.
To produce a fixed-length sequence compatible with UNETR, which requires hidden states from all transformer blocks, we average the modality-specific representations at each spatial location and block.
Formally, given \(\mathrm{v}_{m}^j \in \mathbb{R}^d\) a specific hidden state at location \( j \) for modality \( m \), the aggregated hidden state is defined as
\(
\mathrm{v}^j = \frac{1}{\mid M \mid} \sum_{m \in M} \mathrm{v}_{m}^j
\).
This produces a consistent sequence of tokens, independent of the modality configuration.
All models are fine-tuned for 70 epochs using the AdamW optimizer with a batch size of 2. 
The learning rate is initialized at \(5 \times 10^{-4}\) with a weight decay of 0.05.
Training begins with a 10-epoch warm-up phase, followed by a cosine decay schedule.
\begin{table}[ht]
\caption{Segmentation results comparing models trained from scratch, fine-tuned after conventional
pre-training (separate pre-training for each modality combination), and fine-tuned after BM-MAE pre-
training (one single pre-training).
*: \(p < 0.05\) by Wilcoxon signed-rank test for pairwise comparison with our method.}
\label{tab:seg}
\resizebox{\textwidth}{!}{
\begin{tabular}{|cccc|cccccccccccc|}
\hline
\multicolumn{4}{|c|}{Modalities}                                                                              & \multicolumn{12}{c|}{Dice ($\uparrow$)}                                                                                                                                                                                  \\ \cline{5-16} 
                          &                           &                           &                           & \multicolumn{4}{c|}{Enhancing Tumor}                                           & \multicolumn{4}{c|}{Tumor Core}                                       & \multicolumn{4}{c|}{Whole Tumor}                                \\
F                         & T1c                       & T1                        & T2                        & Scratch & SimCLR         & MAE            & \multicolumn{1}{c|}{BM-MAE}         & Scratch & SimCLR         & MAE   & \multicolumn{1}{c|}{BM-MAE}         & Scratch       & SimCLR         & MAE            & BM-MAE         \\
                          &                           &                           &                           & \cite{hatamizadeh_unetr_2022}       & \cite{ali_self-supervised_2021}              & \cite{10230477}              & \multicolumn{1}{c|}{(ours)}        & \cite{hatamizadeh_unetr_2022}       & \cite{ali_self-supervised_2021}              & \cite{10230477}     & \multicolumn{1}{c|}{(ours)}        & \cite{hatamizadeh_unetr_2022}             & \cite{ali_self-supervised_2021}              & \cite{10230477}              & (ours)        \\ \hline \hline
                          \rowcolor{gray!10}
\checkmark & $\times$                  & $\times$                  & $\times$                  & 43.4*   & \textbf{45.2*} & 43.3           & \multicolumn{1}{c|}{44.4}          & 59.1*   & \textbf{62.1*} & 59.6  & \multicolumn{1}{c|}{60.7}          & 90.3          & \textbf{91.2*} & 90.2           & 90.8          \\ \rowcolor{gray!10}
$\times$                  & \checkmark & $\times$                  & $\times$                  & 78.5*   & 78.8*          & \textbf{80.0}  & \multicolumn{1}{c|}{79.5}          & 81.0*   & 82.4*          & 83.0  & \multicolumn{1}{c|}{\textbf{83.2}} & 73.0*         & 74.2*          & 76.2           & \textbf{76.7} \\ \rowcolor{gray!10}
$\times$                  & $\times$                  & \checkmark & $\times$                  & 34.2*   & 39.1           & \textbf{40.0}  & \multicolumn{1}{c|}{39.7}          & 50.1*   & 55.6*          & 57.2  & \multicolumn{1}{c|}{\textbf{57.3}} & 70.0*         & 72.5*          & 73.6           & \textbf{73.8} \\ \rowcolor{gray!10}
$\times$                  & $\times$                  & $\times$                  & \checkmark & 42.7*   & 43.4*          & 44.1           & \multicolumn{1}{c|}{\textbf{45.4}} & 56.1*   & 56.7*          & 58.9  & \multicolumn{1}{c|}{\textbf{60.0}} & 83.9*         & 83.9*          & 84.4*          & \textbf{85.0} \\
\checkmark & \checkmark & $\times$                  & $\times$                  & 81.7*   & 82.2*          & 82.8*          & \multicolumn{1}{c|}{\textbf{84.1}} & 83.7*   & 84.6*          & 85.1* & \multicolumn{1}{c|}{\textbf{86.2}} & 92.0*         & 92.2*          & 92.8           & \textbf{92.9} \\
\checkmark & $\times$                  & \checkmark & $\times$                  & 46.5*   & 47.8           & 46.5*          & \multicolumn{1}{c|}{\textbf{48.0}} & 62.1*   & 64.8           & 63.8* & \multicolumn{1}{c|}{\textbf{65.0}} & 91.7*         & 92.0           & \textbf{92.3}  & 91.9          \\
\checkmark & $\times$                  & $\times$                  & \checkmark & 46.8*   & \textbf{48.9*} & 47.9           & \multicolumn{1}{c|}{47.5}          & 62.1*   & \textbf{64.8*} & 63.7  & \multicolumn{1}{c|}{63.3}          & 91.5*         & \textbf{92.5}  & 92.3           & 92.1          \\
$\times$                  & \checkmark & \checkmark & $\times$                  & 80.1*   & 80.5*          & 80.7           & \multicolumn{1}{c|}{\textbf{80.8}} & 82.1*   & 82.8*          & 82.9  & \multicolumn{1}{c|}{\textbf{83.7}} & 75.7*         & 76.8*          & \textbf{79.6}  & 79.3          \\
$\times$                  & \checkmark & $\times$                  & \checkmark & 82.2*   & 82.0*          & 82.5           & \multicolumn{1}{c|}{\textbf{83.0}} & 84.1*   & 84.2*          & 85.1* & \multicolumn{1}{c|}{\textbf{85.9}} & 87.5*         & 87.3*          & 88.5           & \textbf{88.8} \\
$\times$                  & $\times$                  & \checkmark & \checkmark & 44.9*   & 48.0           & 47.6           & \multicolumn{1}{c|}{\textbf{48.6}} & 59.8*   & 63.3           & 62.9* & \multicolumn{1}{c|}{\textbf{64.2}} & 86.1*         & 87.3           & \textbf{87.5}  & \textbf{87.5} \\ \rowcolor{gray!10}
\checkmark & \checkmark & \checkmark & $\times$                  & 83.6*   & 84.2           & 83.9           & \multicolumn{1}{c|}{\textbf{84.4}} & 85.2*   & \textbf{86.5}  & 86.3* & \multicolumn{1}{c|}{86.4}          & 92.5*         & 93.2*          & \textbf{93.3}  & 93.0          \\ \rowcolor{gray!10}
\checkmark & \checkmark & $\times$                  & \checkmark & 82.8*   & 83.1*          & \textbf{84.0*} & \multicolumn{1}{c|}{\textbf{84.0}} & 85.3*   & 85.3*          & 86.6* & \multicolumn{1}{c|}{\textbf{87.0}} & 93.1*         & 93.6*          & \textbf{94.0}  & 93.5          \\ \rowcolor{gray!10}
\checkmark & $\times$                  & \checkmark & \checkmark & 48.5    & \textbf{50.0*} & 48.9           & \multicolumn{1}{c|}{48.7}          & 64.0*   & \textbf{66.2}  & 65.0  & \multicolumn{1}{c|}{64.9}          & \textbf{92.6} & 92.5           & \textbf{92.6}  & \textbf{92.6} \\ \rowcolor{gray!10}
$\times$                  & \checkmark & \checkmark & \checkmark & 82.8*   & 82.6*          & 83.2           & \multicolumn{1}{c|}{\textbf{83.3}} & 85.2*   & 85.0*          & 85.6  & \multicolumn{1}{c|}{\textbf{86.4}} & 88.4*         & 88.1*          & \textbf{89.1}  & 88.8          \\ 
\checkmark & \checkmark & \checkmark & \checkmark & 83.7*   & 84.4           & \textbf{84.8}  & \multicolumn{1}{c|}{\textbf{84.8}} & 85.5*   & 86.6           & 86.8* & \multicolumn{1}{c|}{\textbf{87.0}} & 93.4          & 94.0           & \textbf{94.1*} & 93.7          \\ \hline \hline
\rowcolor{gray!30}
\multicolumn{4}{|c|}{Average}                                                                                 & 64.2    & 65.3           & 65.3           & \multicolumn{1}{c|}{\textbf{65.7}} & 72.4    & 74.1           & 74.2  & \multicolumn{1}{c|}{\textbf{74.8}} & 86.8          & 87.4           & \textbf{88.0}  & \textbf{88.0} \\ \hline
\end{tabular}
}
\end{table}

\paragraph{Results.}
Table~\ref{tab:seg} outlines that using BM-MAE consistently enhances Dice scores across the three classes compared to training from scratch.
On average, we observe Dice score increases of 2.3\% for ET, 3.3\% for TC, and 1.4\% for WT.
These improvements are statistically significant in 42 out of 45 cases, as determined by the Wilcoxon signed-rank test.
Compared to other baselines, we note that, although rarely statistically significant (\(p > 0.05\)), standard MAE pre-training shows competitive performance on the WT label, outperforming BM-MAE in some scenarios. 
When averaged across all modality combinations, the WT results for MAE are on par with those of BM-MAE.
On more challenging subregions, namely ET and TC, our approach outperforms other pre-training strategies across various modality combinations, and consistently achieves better average performance.
This supports the idea that pre-training the encoder on the full set of modalities enables it to learn shared representations enriched with cross-modal interactions.
Importantly, these representations retain their utility even when fine-tuning is performed on incomplete modality subsets, as the pre-trained weights implicitly encode information from the missing modalities.
Hence, BM-MAE offers a strong and versatile backbone for subregion segmentation, requiring only a single pre-training phase to handle diverse modality combinations.

\subsection{Downstream Subtyping Classification}
\label{subsec:subtype}
\paragraph{Subtyping LGG vs GBM.}
We further evaluate BM-MAE in a more conventional fine-tuning setting. 
Specifically, we use the model pre-trained without the \textbf{TCGA subset}, and then assess its ability to distinguish between GBM and LGG patients on this same subset. 
This setup allows the model to be evaluated on a classification task using an independent dataset.
Moreover, subtyping remains a clinically relevant task because it guides treatment decisions and is strongly associated with patient prognosis.

\paragraph{Implementation details.}
We use the output of the \texttt{<cls>} token from the ViT as a global representation, followed by a linear layer producing a single scalar output.
We compare four approaches: training from scratch, fine-tuning after SimCLR pre-training, fine-tuning after MAE pre-training, and fine-tuning after BM-MAE pre-training.
All models are trained for 10 epochs using the AdamW optimizer with a batch size of 2. 
The learning rate is initialized at \( 1 \times 10^{-4} \) with a weight decay of 0.05.
Training begins with a 5-epoch warm-up phase, followed by a cosine decay schedule.
We conduct a 5-fold stratified cross-validation aggregate results as mean \( \pm \) standard deviation.

\paragraph{Results.}
Table~\ref{tab:subtype} shows that BM-MAE pre-training consistently outperforms training from scratch across all modality combinations in terms of AUC and AP. 
More specifically, BM-MAE weights yields an average improvement of 21.8\% in AUC and 13.6\% in AP compared to random initialization. 
For the baselines, SimCLR-based pre-training performs markedly worse, with results ranging from acceptable to substantially below training from scratch, and in some cases even underperforming random predictions. 
We hypothesize that this degradation arises from the reliance on augmentations in SimCLR, which can be semantically detrimental.
On the other hand, MAE achieve strong overall performance, but on average fall slightly short of BM-MAE, while still surpassing it in certain modality scenarios but require a separate pre-training for each combination.
We observe that the T1c modality is particularly important for accurately subtyping glioma into LGG and GBM, since LGG typically lack contrast enhancement.
Training from scratch without T1c leads to a sharp drop in performance (AUC: 79.9 → 58.9, AP: 86.6 → 71.9) compared to combinations that include T1c.
BM-MAE also shows a performance drop without T1c (AUC: 91.1 → 78.9, AP: 94.2 → 86.3), but the results remain substantially higher.
This is clinically relevant, as it demonstrates the ability to make accurate predictions even without contrast injection.
In this setting, BM-MAE achieves higher AUC than MAE in five modality scenarios, matches it in one, and is marginally lower in one.
These results further highlight the robustness conferred by pre-training on all modalities, enabling the models to leverage transferable cross-modal representations even when some modalities are missing.

\tabcolsep=0.15cm
\begin{table}[!ht]
\centering
\caption{Brain subtyping performance (LGG vs GBM). Results are reported as AUC (\%) and AP (\%) across modality combinations.}
\label{tab:subtype}
\resizebox{\textwidth}{!}{
\begin{tabular}{|cccc|cccc|cccc|}
\hline
\multicolumn{4}{|c|}{Modalities} & \multicolumn{4}{c|}{AUC ($\uparrow$)} & \multicolumn{4}{c|}{AP ($\uparrow$)}    \\ 
F                         & T1c                       & T1                        & T2                        & Scratch & SimCLR         & MAE            & \multicolumn{1}{c|}{BM-MAE}         & Scratch & SIMCLR         & MAE   & \multicolumn{1}{c|}{BM-MAE}        \\
                          &                           &                           &                           &      & \cite{ali_self-supervised_2021}              & \cite{10230477}              & \multicolumn{1}{c|}{(ours)}        &      & \cite{ali_self-supervised_2021}              & \cite{10230477}     & \multicolumn{1}{c|}{(ours)} \\ \hline \hline
\rowcolor{gray!10}
     \checkmark     &     $\times$    &   $\times$   &  $\times$    & 64.2 $\pm$ 7.1 & 47.6 $\pm$ 5.7 & \textbf{73.9 $\pm$ 8.3} & \textbf{73.9 $\pm$ 5.4} & 77.6 $\pm$ 5.2 & 61.7 $\pm$ 2.3 & \textbf{82.5 $\pm$ 6.9} & 82.3 $\pm$ 5.0 \\ \rowcolor{gray!10}
    $\times$     &   \checkmark      &  $\times$    &   $\times$   & 82.4 $\pm$ 8.1 & 62.7 $\pm$ 7.8 & 90.7 $\pm$ 4.8 & \textbf{91.6 $\pm$ 3.4} & 88.0 $\pm$ 6.8 & 72.4 $\pm$ 4.9 & 94.1 $\pm$ 3.3 & \textbf{94.9 $\pm$ 1.6}  \\ \rowcolor{gray!10}
     $\times$    &    $\times$     &  \checkmark    &  $\times$    &  63.5 $\pm$ 12.3 & 55.5 $\pm$ 12.5 & 77.1 $\pm$ 9.4 & \textbf{83.8 $\pm$ 2.5} & 73.3 $\pm$ 8.4 & 69.0 $\pm$ 10.6 & 84.9 $\pm$ 6.3 & \textbf{90.4 $\pm$ 2.7}  \\ \rowcolor{gray!10}
    $\times$     &   $\times$      &    $\times$  &  \checkmark    &  49.5 $\pm$ 15.6 & 48.3 $\pm$ 8.8 & 74.1 $\pm$ 6.4 & \textbf{75.1 $\pm$ 6.2} & 61.5 $\pm$ 8.4 & 60.7 $\pm$ 4.4 & \textbf{83.6 $\pm$ 3.7} & 82.1 $\pm$ 5.3 \\
    \checkmark     &     \checkmark    &   $\times$   &  $\times$    &   80.9 $\pm$ 6.2 & 66.9 $\pm$ 8.8 & 87.7 $\pm$ 3.8 & \textbf{92.5 $\pm$ 2.5}& 87.6 $\pm$ 6.7 & 79.0 $\pm$ 6.3 & 92.6 $\pm$ 2.5 & \textbf{95.7 $\pm$ 1.2}  \\
  \checkmark       &    $\times$     &   \checkmark   &    $\times$  & 62.9 $\pm$ 7.4 & 68.2 $\pm$ 4.7 & 73.9 $\pm$ 7.0 & \textbf{83.6 $\pm$ 6.9}& 77.1 $\pm$ 5.4 & 79.4 $\pm$ 4.4 & 81.8 $\pm$ 4.1 & \textbf{90.4 $\pm$ 4.1}\\
    \checkmark     &    $\times$     &    $\times$  &   \checkmark   &  57.9 $\pm$ 6.4 & 62.3 $\pm$ 10.9 & \textbf{78.4 $\pm$ 5.9} & 76.8 $\pm$ 7.6& 72.8 $\pm$ 4.2 & 72.6 $\pm$ 8.3 & \textbf{87.0 $\pm$ 4.5} & 84.4 $\pm$ 4.4  \\
    $\times$     &  \checkmark       &  \checkmark    &  $\times$    &   82.8 $\pm$ 7.0 & 59.5 $\pm$ 9.5 & \textbf{91.4 $\pm$ 5.1} & 91.2 $\pm$ 4.9& 89.6 $\pm$ 4.1 & 69.0 $\pm$ 6.7 & \textbf{95.1 $\pm$ 3.2} & 93.8 $\pm$ 4.2 \\
   $\times$      &   \checkmark      &   $\times$   &  \checkmark    & 80.5 $\pm$ 5.3 & 56.7 $\pm$ 18.1 & 90.5 $\pm$ 4.4 & \textbf{92.4 $\pm$ 3.6}& 87.7 $\pm$ 5.1 & 70.9 $\pm$ 12.4 & 92.6 $\pm$ 4.3 & \textbf{95.3 $\pm$ 2.2} \\
    $\times$     &    $\times$     &  \checkmark     &  \checkmark     & 58.4 $\pm$ 13.1 & 51.1 $\pm$ 7.4 & 79.3 $\pm$ 6.0 & \textbf{80.1 $\pm$ 6.5}& 70.0 $\pm$ 5.6 & 63.8 $\pm$ 6.4 & 86.2 $\pm$ 3.5 & \textbf{87.1 $\pm$ 4.5} \\ \rowcolor{gray!10}
     \checkmark    &   \checkmark      &  \checkmark    &  $\times$    &  82.7 $\pm$ 5.9 & 63.2 $\pm$ 6.3 & \textbf{90.0 $\pm$ 3.2} & 89.4 $\pm$ 2.1& 87.8 $\pm$ 6.5 & 73.6 $\pm$ 6.7 & \textbf{94.4 $\pm$ 1.8} & 93.1 $\pm$ 2.3 \\ \rowcolor{gray!10}
     \checkmark    &    \checkmark     &  $\times$    &  \checkmark    &  74.5 $\pm$ 4.5 & 66.8 $\pm$ 10.8 & \textbf{92.1 $\pm$ 1.4} & 89.8 $\pm$ 3.7& 83.2 $\pm$ 3.5 & 77.7 $\pm$ 10.1 & \textbf{95.3 $\pm$ 0.7} & 93.2 $\pm$ 2.7  \\ \rowcolor{gray!10}
     \checkmark    &   $\times$      &  \checkmark    &  \checkmark    & 55.7 $\pm$ 9.3 & 65.2 $\pm$ 11.9 & 77.1 $\pm$ 6.9 & \textbf{79.2 $\pm$ 8.6}& 70.8 $\pm$ 5.3 & 76.7 $\pm$ 9.9 & 85.5 $\pm$ 4.5 & \textbf{87.4 $\pm$ 6.5} \\ \rowcolor{gray!10}
     $\times$    &   \checkmark      &   \checkmark   & \checkmark     &  80.3 $\pm$ 6.1 & 56.2 $\pm$ 7.7 & 90.9 $\pm$ 3.2 & \textbf{91.7 $\pm$ 3.9} & 85.8 $\pm$ 5.1 & 65.2 $\pm$ 5.5 & 94.7 $\pm$ 2.1 & \textbf{94.9 $\pm$ 2.3}  \\
    \checkmark     &     \checkmark    &   \checkmark   &  \checkmark    &  74.7 $\pm$ 3.9 & 58.3 $\pm$ 22.3 & \textbf{92.6 $\pm$ 1.7} & 89.8 $\pm$ 4.1& 82.9 $\pm$ 4.0 & 72.3 $\pm$ 16.1 & \textbf{95.6 $\pm$ 1.1} & 92.6 $\pm$ 3.9 \\ \hline \hline
    \rowcolor{gray!30}
\multicolumn{4}{|c|}{Average}    &   70.1 $\pm$ 11.2           &  59.2 $\pm$ 6.4         &  84.0 $\pm$ 7.4     &  \textbf{85.4 $\pm$ 6.6}      & 79.7 $\pm$ 7.8 &   70.9 $\pm$ 5.8 & 89.7 $\pm$ 5.1 & \textbf{90.5 $\pm$ 4.6}  \\ \hline
\end{tabular}
}
\end{table}
\subsection{Downstream Survival Analysis} 
\paragraph{General information.}
We use the \textbf{TCGA subset}, which provides overall survival (OS) data, to evaluate survival analysis as an additional clinical task.
In the latter, we aim to model the time \( T \in \mathbb{R}^+ \) until an event of interest occurs, often under right-censoring. 
The survival dataset consists of triplets \((T_i, \delta_i, \mathcal{X}_i)\), with \(\mathcal{X}_i\) the input features (e.g., multimodal MRI data for patient \( i \)), \(T_i\) the observed time and \(\delta_i \) indicates whether the event is observed (\(\delta_i = 1\)) or censored (\(\delta_i = 0\)). 
The objective is to use a neural network \( \phi_\theta (\mathcal{X}_i)\) to estimate the hazard function,  expressed in continuous time as:
\[
h(t \mid \mathcal{X}) = \lim_{\Delta t \to 0} \frac{P(t \leq T < t + \Delta t \mid T \geq t, \mathcal{X})}{\Delta t}
\]
that represents the instantaneous event rate at time \(t\), given that the event has not occurred yet.
In a discrete-time setting, the hazard simplifies to \(h(t \mid \mathcal{X})= P(T=t \mid T \geq t, \mathcal{X}) \).
Hence, to make the problem tractable for deep learning, time is discretized into \(K\) non-overlapping intervals \(( t_1, \dots, t_K)\).
The neural network \(\phi_\theta(\mathcal{X}_i)\) outputs logits \({a}_i \in \mathbb{R}^K\), with each hazard modeled by
\(
h_k(\mathcal{X}_i) = \sigma(a_{ik}),
\)
where \(\sigma(\cdot)\) is the sigmoid function.
Finally, the negative log-likelihood over the \( N \) patients is given by:
\[
\ell_{hazard}(\theta) = - \sum_{i=1}^N \left[ \delta_i \log\left(h_{k(i)}(\mathcal{X}_i)\right) + \sum_{j=1}^{k(i) - \delta_i} \log\left(1 - h_j(\mathcal{X}_i)\right) \right]
\]
where \(k(i)\) is the interval containing \(T_i\).
Learning this hazard function enables individual risk estimation, which is essential for stratifying glioma patients and predicting disease progression from MRI.

\paragraph{Implementation details.}
Models are trained for 5 epochs to mitigate overfitting, which is common in survival analysis.
We partition the time axis into \( K = 10 \) intervals based on empirical quantiles, promoting a balanced distribution of events across intervals.
We use the NLLLogisticHazard loss implemented in Pycox \cite{kvamme_continuous_2021}, which extends the approach proposed in \cite{gensheimer_scalable_2019}.
Models are evaluated using the concordance index (C-index) \cite{antolini_time-dependent_2005}.
The C-index assesses the model's ability to correctly rank survival times, reflecting the probability that predicted risks align with actual event orderings.
All other hyperparameters and system configurations are similar to those used in the subtyping experiments described in Subsection~\ref{subsec:subtype} to ensure consistency.

\paragraph{Results.}
Table~\ref{tab:surv} reports the results for the survival prediction task. Models initialized with BM-MAE pre-trained weights consistently outperform those trained from scratch across all modality combinations. On average, this corresponds to a 23.3\% improvement in C-index.
Among individual modalities, T1c emerges as particularly informative for survival prediction.
For the SimCLR baseline, we observe similar variability across modality combinations, with performance fluctuating substantially. 
MAE pre-training attains strong results overall and, on average, slightly surpasses BM-MAE. Interestingly, our model tends to perform best either with a single modality or with many modalities (three or four), while consistently underperforming with exactly two. 
Given the inherent instability of survival prediction on relatively small datasets, we hypothesize that these fluctuations reflect the variability of training rather than fundamental differences between methods. 
Importantly, BM-MAE still demonstrates strong relevance in this setting, as it achieves competitive survival performance with a single pre-trained model, whereas the other approaches require 15 different models.
\tabcolsep=0.22cm
\begin{table}[!ht]
\centering
\caption{Survival prognosis performance. Results are reported as concordance index (C-index) across modality combinations.}
\label{tab:surv}
\scriptsize
\begin{tabular}{|cccc|cccc|}
\hline
\multicolumn{4}{|c|}{Modalities} & \multicolumn{4}{c|}{C-index ($\uparrow$)} \\
F                         & T1c                       & T1                        & T2                        & Scratch & SIMCLR         & MAE            & \multicolumn{1}{c|}{BM-MAE}    \\
                          &                           &                           &                           &      & \cite{ali_self-supervised_2021}              & \cite{10230477}              & \multicolumn{1}{c|}{(ours)} \\ \hline \hline
\rowcolor{gray!10}
     \checkmark     &     $\times$    &   $\times$   &  $\times$    &  49.5 $\pm$ 3.9 & 48.3 $\pm$ 2.4 & 65.7 $\pm$ 4.0 & \textbf{67.6 $\pm$ 7.6} \\ \rowcolor{gray!10}
    $\times$     &   \checkmark      &  $\times$    &   $\times$   & 65.2 $\pm$ 7.2 & 53.3 $\pm$ 9.4 & 67.8 $\pm$ 2.3 & \textbf{69.1 $\pm$ 3.1} \\ \rowcolor{gray!10}
     $\times$    &    $\times$     &  \checkmark    &  $\times$    & 49.8 $\pm$ 6.3 & 53.0 $\pm$ 8.3 & 60.0 $\pm$ 4.3 & \textbf{63.8 $\pm$ 7.6} \\ \rowcolor{gray!10}
    $\times$     &   $\times$      &    $\times$  &  \checkmark    & 49.0 $\pm$ 7.9 & 52.0 $\pm$ 6.1 & 63.6 $\pm$ 5.9 & \textbf{65.9 $\pm$ 4.1} \\
    \checkmark     &     \checkmark    &   $\times$   &  $\times$    &  61.5 $\pm$ 8.0 & 47.4 $\pm$ 8.8 & \textbf{70.9 $\pm$ 4.5} & 69.4 $\pm$ 6.2 \\
  \checkmark       &    $\times$     &   \checkmark   &    $\times$  &  50.7 $\pm$ 7.1 & 52.6 $\pm$ 8.0 & \textbf{67.8 $\pm$ 5.9} & 65.9 $\pm$ 6.3 \\
    \checkmark     &    $\times$     &    $\times$  &   \checkmark   &    46.4 $\pm$ 5.0 & 58.7 $\pm$ 8.0 & \textbf{69.8 $\pm$ 4.9} & 65.2 $\pm$ 8.3 \\
    $\times$     &  \checkmark       &  \checkmark    &  $\times$    &   62.2 $\pm$ 9.0 & 42.1 $\pm$ 5.3 & \textbf{68.1 $\pm$ 2.8} & 67.3 $\pm$ 6.0  \\
   $\times$      &   \checkmark      &   $\times$   &  \checkmark    &  59.8 $\pm$ 8.6 & 42.0 $\pm$ 9.1 & \textbf{70.1 $\pm$ 6.2} & 66.2 $\pm$ 7.2\\
    $\times$     &    $\times$     &  \checkmark     &  \checkmark     &  48.1 $\pm$ 7.7 & 46.9 $\pm$ 4.5 & \textbf{64.9 $\pm$ 2.5} & 64.5 $\pm$ 8.2\\ \rowcolor{gray!10}
     \checkmark    &   \checkmark      &  \checkmark    &  $\times$    &  59.5 $\pm$ 7.2 & 51.0 $\pm$ 7.5 & \textbf{73.4 $\pm$ 2.5} & 68.3 $\pm$ 2.2 \\ \rowcolor{gray!10}
     \checkmark    &    \checkmark     &  $\times$    &  \checkmark    &  55.3 $\pm$ 7.4 & 58.5 $\pm$ 6.9 & \textbf{71.8 $\pm$ 3.4} & 67.2 $\pm$ 2.1\\ \rowcolor{gray!10}
     \checkmark    &   $\times$      &  \checkmark    &  \checkmark    &  47.2 $\pm$ 4.8 & 57.9 $\pm$ 3.2 & 66.2 $\pm$ 6.1 & \textbf{67.9 $\pm$ 5.1} \\ \rowcolor{gray!10}
     $\times$    &   \checkmark      &   \checkmark   & \checkmark     &  58.4 $\pm$ 5.7 & 47.6 $\pm$ 2.8 & 68.4 $\pm$ 2.5 & \textbf{69.5 $\pm$ 4.7}\\
    \checkmark     &     \checkmark    &   \checkmark   &  \checkmark    & 59.5 $\pm$ 5.7 & 44.3 $\pm$ 7.0 & 71.6 $\pm$ 6.3 & \textbf{71.9 $\pm$ 2.0} \\ \hline \hline
    \rowcolor{gray!30}
\multicolumn{4}{|c|}{Average}    &        54.6 $\pm$ 6.2      &   50.4 $\pm$ 5.3    &   \textbf{68.0 $\pm$ 3.4} & 67.3 $\pm$ 2.1\\ \hline
\end{tabular}
\end{table}

\subsection{Cross-modal relationships}
\begin{figure*}[!ht]
\centering
\resizebox{\textwidth}{!}{
\includegraphics[scale=0.2]{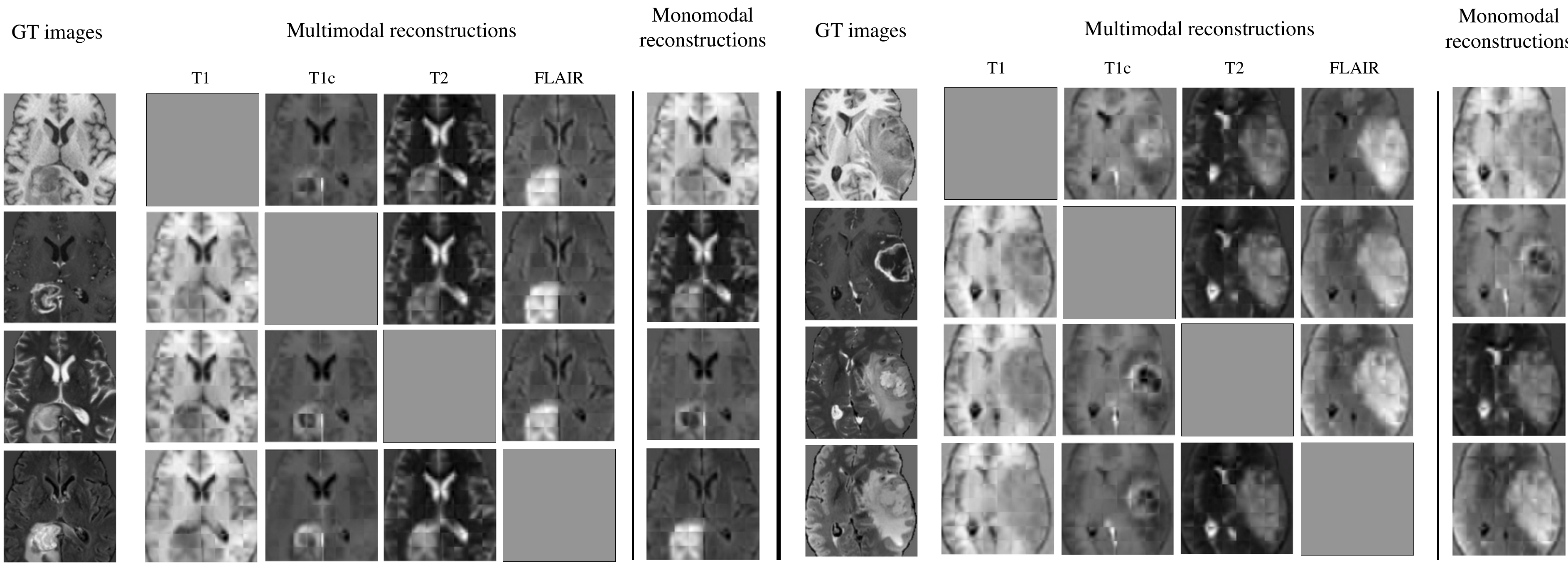}}
\caption{Reconstruction examples. For each of the two samples, from left to right: the original modalities, the reconstruction of three modalities from a single modality, and the reconstruction of each modality from the other three modalities.}
\label{fig:recons}
\end{figure*}

\paragraph{Reconstruction visualization.}
Although BM-MAE uses a shallow decoder to focus on representation learning rather than high-fidelity reconstruction, it still enables flexible cross-modal reconstruction by selecting appropriate masks.
This flexibility supports scenarios such as reconstructing three modalities from one or a single modality from the other three.
Figure~\ref{fig:recons} shows that the model learn to produce realistic reconstructions, preserving key semantic structures despite some blurring and patch artifacts.
Results are particularly impressive, especially given the minimal computation time enabled by parallelization through attention blocks to reconstruct multiple 3D volumes.

\begin{figure*}[!ht]
\centering
\resizebox{\textwidth}{!}{
\includegraphics[scale=0.7]{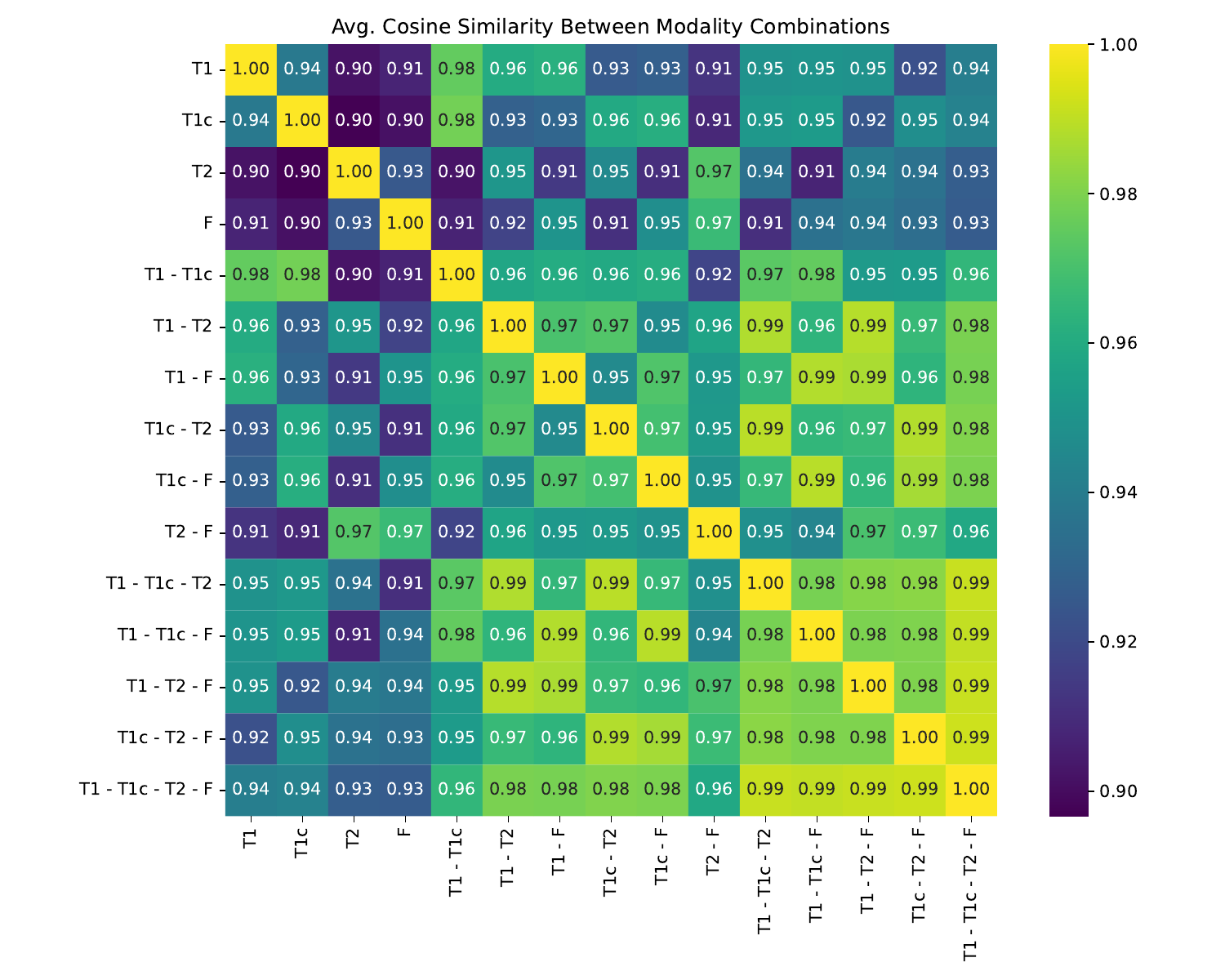}
}
\caption{For each patient, we compute embeddings under all possible modality combinations and evaluate pairwise cosine similarities between them.
This analysis characterizes how consistent the output representations remain when varying the set of input modalities.}
\label{fig:cosine_sim}
\end{figure*}

\paragraph{Cross-modal consistency analysis.}
Consistency of patient embeddings across different subsets of input modalities provides valuable insight into the nature of the learned representations.
To evaluate this property in BM-MAE, we generate an embedding for each patient under every possible combination of input modalities and compare them by computing pairwise cosine similarities. 
For each pair of modality combinations, we then average the cosine similarity across patients, yielding a square matrix whose size corresponds to the number of modality combinations, as illustrated in Figure~\ref{fig:cosine_sim}. 
This summarizes the stability of patient representations under varying input conditions.
Overall, embeddings exhibit high cross-modal similarity, with average values exceeding 0.90 across all comparisons. 
The lowest similarities occur between combinations that do not share any modalities, reflecting the greater challenge of aligning disjoint data types while still demonstrating substantial coherence.
Importantly, embeddings become progressively closer to the full-modal representation as more modalities are included. 
On average, cosine similarity reaches 0.94 when using a single modality, 0.97 with two modalities, and 0.99 with three modalities, compared to the full-modal case.
These results are particularly noteworthy given that BM-MAE does not rely on any contrastive or alignment objective: the observed consistency emerges purely from the masked reconstruction pre-training. 
This highlights the model’s ability to naturally integrate heterogeneous inputs into a shared latent space, supporting robustness to missing modalities.
\section{Conclusion}
In this work, we introduce BM-MAE, a multimodal pre-training strategy for anatomical brain tumor MRI analysis that effectively handles missing modalities.
Our approach enables a single model, pre-trained once on the full set of available modalities, to serve as a universal backbone for downstream tasks, regardless of which subset of modalities is used.
This is particularly relevant in the medical domain, where data availability is often incomplete and specialized datasets typically provide only a subset of imaging modalities.
BM-MAE enables the integration of information from all modalities during pre-training, allowing the resulting representations to remain informative even when only partial inputs are available during fine-tuning.
We demonstrate that this approach yields significant improvements across several clinically important tasks, including glioma subtyping, tumor subregion segmentation and survival analysis even when key modalities are missing.
The model consistently outperforms training from scratch and achieves performance that matches or exceeds that of conventional pre-training approaches that require retraining for each subset of modalities.
We hope that these distinctive capabilities will encourage its adoption within the clinical research community and catalyze further studies.
Ultimately, this could pave the way towards advancing personalized medicine for glioma patients.

\bibliographystyle{unsrt}  
\bibliography{references}
\clearpage
\appendix
\renewcommand{\thefigure}{S\arabic{figure}}
\section*{Supplementary Material}
\addcontentsline{toc}{section}{Supplementary Material}
\section{Reconstruction examples from masked inputs}
\begin{figure*}[!ht]
\resizebox{\textwidth}{!}{
\includegraphics[scale=0.26]{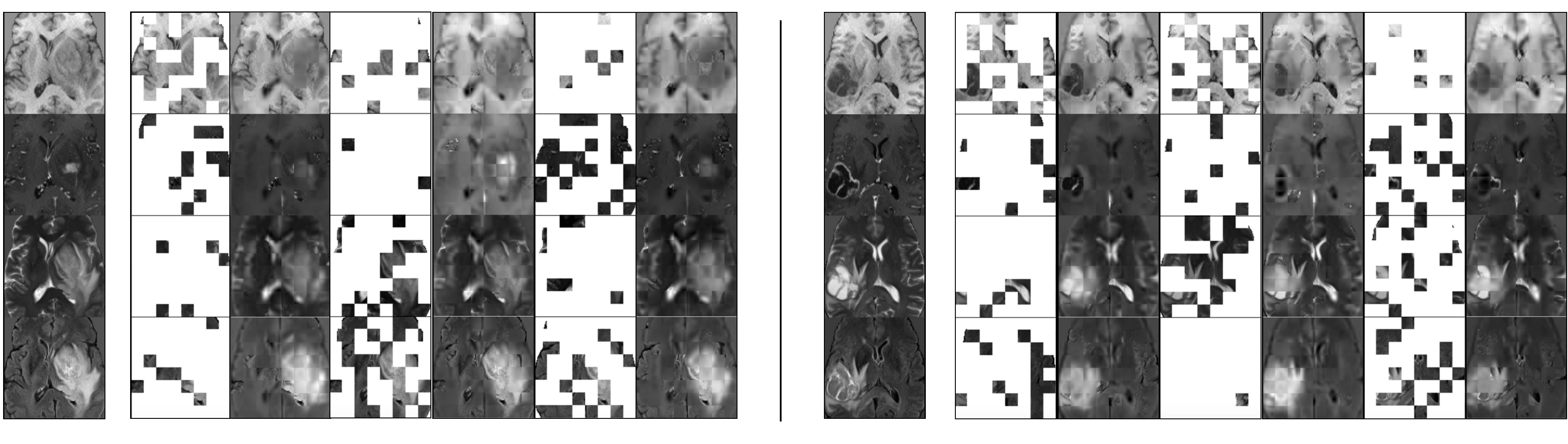}}
\caption{Examples of reconstructions from Dirichlet-based mask sampling. For each sample, we display the original image along with several pairs of masked inputs and reconstructions, where predictions are overlaid with visible patches, as loss is applied only to masked regions.}
\label{fig:mask_sampling}
\end{figure*}
\end{document}